\documentclass[runningheads]{llncs}
\usepackage[T1]{fontenc}
\usepackage{booktabs}
\usepackage{graphicx,verbatim}
\usepackage{amsmath, amsfonts}
\usepackage{multirow}
\usepackage{hyperref}
\usepackage{siunitx}
\usepackage{makecell}
\usepackage{marvosym}
\usepackage{orcidlink}
\usepackage{tabularx}
\usepackage{array}

\renewcommand{\orcidID}[1]{\,\orcidlink{#1}} 
\begin{document}
\title{Do Segmentation Encoders Learn Diagnostic Representations?  A Skin-Lesion Classification Study}
\titlerunning{Do Segmentation Encoders Learn Diagnostic Representations?}
%

\author{Javier Guinea-Pérez\inst{1}\orcidID{0009-0006-6656-5612}\textsuperscript{(\Letter)}\thanks{These authors contributed equally to this work.} \and
Alessandro Ceressi \inst{1}\orcidID{0009-0002-9857-9518}\protect\footnotemark[1] \and
Anaida Fernández García
\inst{1}\orcidID{0000-0002-6102-3121}\protect\footnotemark[1] \and
Pablo Perdomo Quinteiro\inst{1}\orcidID{0000-0001-8784-0907}\protect\footnotemark[1] \and
Federico Álvarez\inst{1}\orcidID{0000-0001-7400-9591}}
\authorrunning{J. Guinea-Pérez et al.}
\institute{Grupo de Aplicación de Telecomunicaciones Visuales, Information Processing and Telecommunications Center, Universidad Politécnica de Madrid, 28040 Madrid, Spain\\
\email{javier.guinea.perez@upm.es}}
  
\maketitle              
\begin{abstract}
Promptable segmentation foundation models transfer well across segmentation tasks, but their usefulness as image-level diagnostic representations remains unclear. We evaluate whether SAM2/Hiera-based encoders can serve as frozen embeddings for skin-lesion classification. The original SAM2 encoder and a medically fine-tuned MedSAM2 encoder are compared with DINOv2 across multiple dermatology datasets using a common frozen-feature protocol and lightweight downstream classifiers. SAM2-derived embeddings transfer partially to diagnosis, and the medically fine-tuned MedSAM2 improves over the original SAM2 encoder on average. However, DINOv2 achieves the strongest average macro-F1 and obtains the highest point estimate on all evaluated datasets. These results suggest that medical segmentation fine-tuning can improve representation transfer, but segmentation-oriented pretraining does not necessarily yield the most discriminative embeddings for skin-lesion classification.


\end{abstract}
\section{Introduction}

Image foundation models have reshaped computer vision by allowing models trained at large scale to be reused for a wide range of downstream applications. In medical imaging, this is a highly attractive proposition; annotated medical data is difficult to obtain and often comes with complex privacy, acquisition and institutional constraints. Promptable segmentation models are an example of this paradigm. Segment Anything Model (SAM) \cite{kirillov_segment_2023} and its developments SAM2 \cite{ravi_sam_2024} and SAM3 \cite{carion_sam_2026} have demonstrated generalization capabilities, and have been adapted and fine-tuned for medical image segmentation numerous times \cite{ali_review_2025}.

The generalization capabilities of these foundation segmentation models have been tested within segmentation and perception-related tasks, such as edge detection, object proposal or text-to-mask. It remains unclear whether the semantic representations within these models are transferable to non-segmentation downstream use. We ask whether the encoder representations actually transfer to a different downstream objective: skin-lesion diagnosis/classification.

Segmentation and diagnostic classification impose different representational requirements. A segmentation encoder is expected to construct representations that capture object boundaries, texture, spatial layout, and foreground-background separation, while a diagnostic classifier must separate clinically meaningful categories, often under class imbalance and subtle inter-class visual similarity. Dermatology provides a useful test case: skin-lesion diagnosis depends on morphology, color, texture, borders, and lesion structure, all visual cues that segmentation encoders might capture. 

In this work, we evaluate whether the encoders of SAM2-based models can serve as general-purpose embeddings for skin-lesion classification. We compare the original SAM2 encoder and a medically fine-tuned SAM2 encoder against a standard representation baseline (DINOv2 \cite{oquab_dinov2_2024}) across several dermatological datasets. To isolate the contribution of the pretrained representation, all image encoders are kept frozen. Classification is performed using lightweight downstream heads: a validation-tuned multilayer perceptron. 

\section{Previous work}

Promptable segmentation foundation models have established a new paradigm for dense visual prediction. SAM introduced a promptable segmentation framework in which points, boxes, masks, and other prompts are used to generate object masks across diverse image distributions \cite{kirillov_segment_2023}. SAM2 extended this formulation from still images to videos by introducing memory-based mask propagation, enabling promptable visual segmentation across time \cite{ravi_sam_2024}. More recently, SAM3 broadened the SAM family from primarily geometry-based prompts to concept-conditioned segmentation, combining text and image prompts for open-vocabulary detection, segmentation, and tracking \cite{carion_sam_2026}. Across these models, generalization is primarily demonstrated through segmentation or segmentation-adjacent perception tasks, including mask generation, object proposal, visual prompting, video object segmentation, open-vocabulary detection, and concept-conditioned mask prediction.

This segmentation-centric formulation has also motivated extensive work on adapting SAM-family models to specialized domains. In medical imaging, the direct application of SAM is limited by the domain gap between training images and clinical data, but fine-tuning and adaptation have been shown to improve segmentation performance across modalities and anatomical targets. MedSAM fine-tunes SAM for medical image segmentation across a broad collection of organs, lesions, and imaging modalities \cite{ma_segment_2024}. Other works adapt SAM through lightweight modules, adapters, or task-specific training for 2D, 3D, and volumetric medical segmentation \cite{ali_review_2025}. MedSAM2 extends this line of work by adapting SAM2 to 3D medical images and videos, using medical-domain fine-tuning to improve segmentation across CT, MRI, PET, ultrasound, and endoscopy data \cite{ma_medsam2_2025}. These works provide strong evidence that SAM models can be transferred to medical segmentation tasks, but their evaluation remains focused on mask quality rather than image-level diagnostic representation.

The success of segmentation foundation models in generalization raises a different question: are their internal visual representations also useful for downstream tasks outside segmentation. Recent work has begun to examine this distinction. Studies probing SAM as a backbone for visual understanding report that SAM and SAM2 encoders are substantially weaker than representation-oriented models such as CLIP \cite{radford_learning_2021} and DINOv2 \cite{oquab_dinov2_2024} on semantic classification benchmarks \cite{espinosa_there_2024}. Related work on SAM-CLIP also argues that different foundation models encode complementary capabilities: SAM provides strong spatial understanding, while CLIP provides stronger semantic understanding \cite{wang_sam-clip_2024}. These findings suggest that the pretraining objective of a foundation model strongly influences the downstream utility of its encoder. Recent approaches also use the frozen SAM image encoder as a feature extractor for medical tasks, adding trainable classification modules to adapt segmentation-derived features to medical classification tasks \cite{gu_adapting_2025}. Broader medical transfer-learning studies have also compared natural-domain foundation models such as DINOv2, MAE, CLIP-like models, and SAM2 for medical image classification across datasets \cite{mansoori_advancements_2025}.

Skin-lesion classification is a particularly relevant test case for image transfer-learning. Dermatological images contain visual cues that overlap with segmentation-relevant features, including lesion boundaries, color distribution, texture, asymmetry, and differential structure \cite{nachbar_abcd_1994}.

\section{Materials and methods}

\begin{figure}
    \centering
    \includegraphics[width=\linewidth]{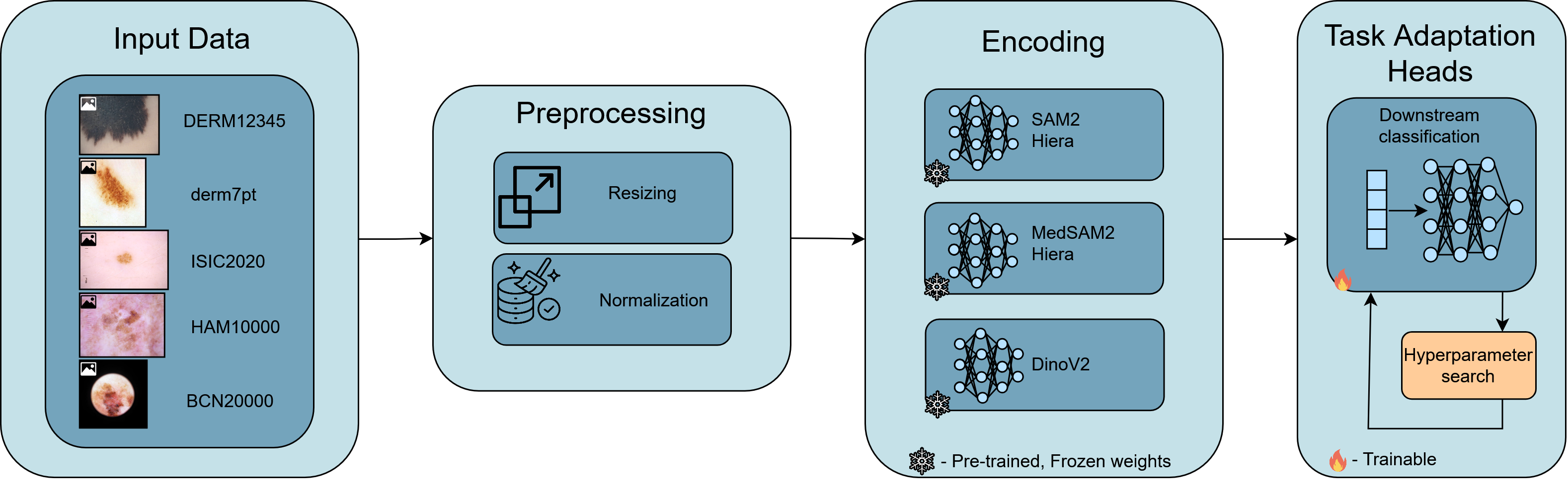}
    \caption{Overview of the frozen-feature evaluation pipeline. Dermatology images are loaded, resized, and normalized using a common preprocessing protocol before being passed through a frozen pretrained image encoder: SAM2/Hiera, MedSAM2/Hiera, or DINOv2. The resulting fixed embeddings are extracted once and used to train a lightweight validation-tuned MLP classifier for skin-lesion diagnosis, while the image encoder remains unchanged throughout downstream training.}
    \label{fig:diagram}
\end{figure}

This study evaluates whether image encoders from segmentation foundation models can be reused as frozen embedding extractors for skin-lesion classification. We compare SAM2/Hiera-based encoders with a non-segmentation visual baseline under a common downstream evaluation protocol. In all experiments, the image encoder is kept frozen and only a lightweight classification head is trained. Figure \ref{fig:diagram} shows a diagram of the pipeline.

\subsection{Encoding}

For each dataset, images are processed using a common embedding extraction pipeline. Images are loaded through a dataset-specific adapter, resized according to the input resolution of the encoder, and normalized using ImageNet statistics \cite{deng_imagenet_2009}. Thus, all encoders use the same dataset loading, channel handling, and normalization strategy, while the spatial resolution is encoder-specific.

The preprocessed images are passed through a frozen pretrained image encoder in evaluation mode. No encoder parameters are updated during downstream training. All embeddings are extracted once and reused across downstream hyperparameter trials.

For the SAM2/Hiera and MedSAM2/Hiera encoders, we use the Hiera trunk and retain the intermediate feature maps returned by the encoder. The image-level embedding is obtained from the final-stage feature map. Specifically, if the last Hiera feature map has shape $B \times C \times h \times w$, we apply global average pooling over the spatial dimensions to obtain a single $B \times C$ embedding per image. For the Hiera-Tiny encoders used in this work, this produces 768-dimensional image embeddings.

For DINOv2, we use the \texttt{vit\_small\_patch14\_dinov2} model from \texttt{timm} with the classification head removed. Image embeddings are extracted from the output of \texttt{forward\_features}. We use the normalized class token when available, and otherwise the first token of the final token sequence. This produces one 384-dimensional image-level embedding per input image.

\begin{table}[t]
\centering
\caption{Frozen image encoders evaluated in this study.}
\label{tab:encoders}
\small
\setlength{\tabcolsep}{5pt}
\renewcommand{\arraystretch}{1.15}

\begin{tabularx}{\linewidth}{@{}l
    >{\raggedright\arraybackslash}X
    c
    >{\raggedright\arraybackslash}X@{}}
\toprule
\textbf{Encoder} & 
\textbf{Backbone} & 
\textbf{Input size} & 
\textbf{Pretraining/adaptation} \\
\midrule
SAM2/Hiera 
    & Hiera-Tiny 
    & $1024 \times 1024$ 
    & SAM2 segmentation pretraining \\
MedSAM2/Hiera 
    & Hiera-Tiny 
    & $1024 \times 1024$ 
    & Medical SAM2 adaptation \\
DINOv2 
    & ViT-S/14, 22.1M params 
    & $518 \times 518$ 
    & DINOv2 self-supervised LVD-142M \\
\bottomrule
\end{tabularx}
\end{table}

\subsection{Downstream classification heads}

The downstream classifier is trained on top of the frozen embeddings. We use lightweight multilayer perceptron heads with tunable depth and hidden dimension. The use of shallow task-specific heads reflects a practical downstream adaptation setting: the representation remains frozen, but the readout function is selected using validation performance. This allows us to test whether segmentation-derived embeddings contain useful diagnostic information that can be recovered by a simple nonlinear classifier. The embedding representation is passed to a shallow classification head consisting of layer normalization, a linear hidden layer, GELU activation, dropout and a final linear output layer. The depth, hidden dimension and dropout rate are selected by validation performance.

Classification heads are trained using cross-entropy loss. To mitigate class imbalance, class weights are included as a tunable option during hyperparameter search. When enabled, class weights are computed from the training-set label frequencies and normalized so that observed classes have comparable average weight. Label smoothing is also treated as a tunable hyperparameter.

The training objective is defined according to the structure of the target labels. For single-label datasets, the model is trained as a multiclass classifier over mutually exclusive diagnostic categories.

\subsection{Hyperparameter optimization}

Downstream-head hyperparameters are selected using validation performance. We use Optuna to optimize the training configuration for each encoder/dataset combination. The search space includes depth, hidden dimension size, dropout rate, learning rate, weight decay, early-stopping patience and class weighting. 

For binary classification tasks, balanced accuracy is used as the validation selection metric. For multi-class tasks, macro-averaged F1 score is used. 

\subsection{Model selection and evaluation}

After hyperparameter optimization, the best validation trial is evaluated once on the held-out test set. We report macro-F1 score as the primary metric. To quantify uncertainty under the fixed test split, we computed 95\% confidence intervals using 5,000 stratified bootstrap resamples of the held-out test set. For pairwise comparisons between encoders, we used paired stratified bootstrap resampling.

\subsection{Datasets}

We conducted the experiment on five dermatology datasets: HAM10000 \cite{tschandl_ham10000_2018}, Derm7pt \cite{kawahara_seven-point_2019}, DERM12345 \cite{yilmaz_derm12345_2024}, ISIC2020 \cite{rotemberg_patient-centric_2021}, and BCN20000 \cite{hernandez-perez_bcn20000_2024}. These datasets were selected to cover different skin-lesion classification tasks based on individual dermatological images. Table~\ref{tab:datasets} summarizes the datasets used in the study. Binary datasets were formulated as malignant/melanoma versus benign/non-melanoma according to each dataset’s diagnostic labels.

For all datasets, we use predefined splits when available. Otherwise, we construct train, validation, and test patient-level partitions (70/15/15\%). All reported experiments use the same split for all encoders within a dataset.

\begin{table}[t]
\centering
\caption{Summary of the dermatology datasets used in this work.}
\label{tab:datasets}
\small
\setlength{\tabcolsep}{8pt}
\begin{tabular}{@{}llrr@{}}
\toprule
\textbf{Dataset} & \textbf{Image type} & \textbf{Size} & \textbf{Classes} \\
\midrule
HAM10000 \cite{tschandl_ham10000_2018}          & Dermoscopic              & 10,015 & 7 \\
Derm7pt \cite{kawahara_seven-point_2019}        & Clinical and dermoscopic & 1,011  & 2 \\
DERM12345 \cite{yilmaz_derm12345_2024}          & Dermoscopic              & 12,345 & 2 \\
BCN20000  \cite{hernandez-perez_bcn20000_2024}  & Dermoscopic              & 17,790 & 2 \\
ISIC2020  \cite{rotemberg_patient-centric_2021}  & Dermoscopic              & 33,127 & 2 \\
\bottomrule
\end{tabular}
\end{table}

\section{Results}

\begin{figure}
    \centering
    \includegraphics[width=.7\linewidth]{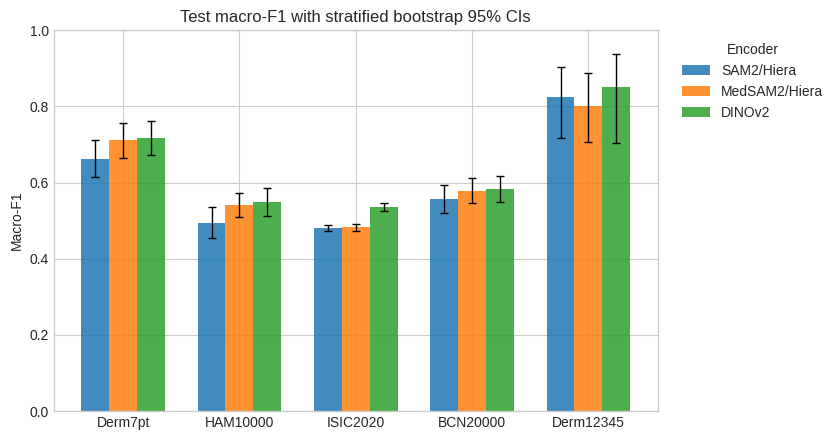}
    \caption{Test macro-F1 of frozen image encoders across dermatology classification datasets. Error bars indicate 95\% confidence intervals estimated from 5,000 stratified bootstrap resamples of the held-out test set.}
    \label{fig:results}
\end{figure}

Tables~\ref{tab:main_results} and ~\ref{tab:pairwise_differences} and Fig.~\ref{fig:results} summarize the classification performance when using the representations from the evaluated frozen encoders across the dermatology datasets. Macro-F1 is used as the primary comparison metric because all tasks involve class imbalance and, in the multi-class setting, overall accuracy may obscure poor minority-class performance. The table reports pairwise differences between encoders and 95\% confidence intervals estimated from 5,000 stratified bootstrap resamples of test sets.

Overall, DINOv2 achieved the highest macro-F1 point estimate on all evaluated datasets and the highest average macro-F1 across datasets. However, the paired bootstrap analysis shows that the robustness of this advantage varies by dataset. Compared with MedSAM2/Hiera, DINOv2 showed a clear advantage on ISIC2020, while the confidence intervals of the paired differences included zero for Derm7pt, HAM10000, BCN20000, and DERM12345. Thus, DINOv2 is the strongest overall point-estimate baseline, but the evidence does not support a uniform or consistently large advantage over MedSAM2 across all datasets.

Medical adaptation of SAM2 produced a modest average improvement over the original SAM2 encoder. MedSAM2/Hiera improved over SAM2/Hiera by an average
of 0.019 macro-F1, with the clearest evidence on HAM10000. The confidence intervals of the paired differences included or reached zero for the remaining datasets, with a negligible point difference on ISIC2020. These results suggest that medical fine-tuning can improve the downstream utility of SAM2-derived frozen representations, but the effect is dataset-dependent.

\begin{table}[t]
\centering
\caption{Bootstrapped test performance of frozen encoders across dermatology datasets. Values report point estimates with 95\% confidence intervals from bootstrap resamples of the held-out test set. Best point estimate per dataset and metric is shown in bold.}
\label{tab:main_results}
\small
\setlength{\tabcolsep}{3pt}
\renewcommand{\arraystretch}{1.10}

\begin{tabular}{@{}llccc@{}}
\toprule
Dataset & Encoder & Macro-F1 & Balanced acc. & AUROC \\
\midrule

Derm7pt
    & SAM2
    & $0.663\,[0.614,\,0.713]$
    & $0.688\,[0.633,\,0.742]$
    & $0.772\,[0.720,\,0.822]$ \\

    & MedSAM2
    & $0.711\,[0.665,\,0.756]$
    & $0.746\,[0.696,\,0.795]$
    & $0.835\,[0.791,\,0.876]$ \\

    & DINOv2
    & $\mathbf{0.718}\,[0.672,\,0.763]$
    & $\mathbf{0.751}\,[0.701,\,0.798]$
    & $\mathbf{0.837}\,[0.791,\,0.879]$ \\

\midrule

HAM10000
    & SAM2
    & $0.495\,[0.455,\,0.537]$
    & $0.562\,[0.517,\,0.609]$
    & $0.918\,[0.905,\,0.930]$ \\

    & MedSAM2
    & $0.541\,[0.509,\,0.574]$
    & $0.647\,[0.602,\,0.690]$
    & $0.924\,[0.906,\,0.938]$ \\

    & DINOv2
    & $\mathbf{0.549}\,[0.514,\,0.585]$
    & $\mathbf{0.671}\,[0.624,\,0.716]$
    & $\mathbf{0.936}\,[0.925,\,0.945]$ \\

\midrule

ISIC2020
    & SAM2
    & $0.481\,[0.472,\,0.489]$
    & $0.750\,[0.701,\,0.795]$
    & $0.843\,[0.800,\,0.881]$ \\

    & MedSAM2
    & $0.483\,[0.474,\,0.491]$
    & $0.776\,[0.731,\,0.818]$
    & $0.856\,[0.813,\,0.895]$ \\

    & DINOv2
    & $\mathbf{0.535}\,[0.525,\,0.546]$
    & $\mathbf{0.816}\,[0.770,\,0.858]$
    & $\mathbf{0.899}\,[0.869,\,0.926]$ \\

\midrule

BCN20000
    & SAM2
    & $0.557\,[0.522,\,0.593]$
    & $0.534\,[0.506,\,0.564]$
    & $0.780\,[0.761,\,0.800]$ \\

    & MedSAM2
    & $0.579\,[0.547,\,0.612]$
    & $\mathbf{0.569}\,[0.538,\,0.601]$
    & $0.801\,[0.783,\,0.819]$ \\

    & DINOv2
    & $\mathbf{0.584}\,[0.550,\,0.619]$
    & $0.562\,[0.536,\,0.592]$
    & $\mathbf{0.841}\,[0.824,\,0.857]$ \\

\midrule

DERM12345
    & SAM2
    & $0.825\,[0.717,\,0.904]$
    & $0.808\,[0.685,\,0.908]$
    & $0.976\,[0.963,\,0.988]$ \\

    & MedSAM2
    & $0.802\,[0.706,\,0.889]$
    & $\mathbf{0.812}\,[0.692,\,0.913]$
    & $\mathbf{0.985}\,[0.975,\,0.993]$ \\

    & DINOv2
    & $\mathbf{0.850}\,[0.705,\,0.938]$
    & $0.799\,[0.678,\,0.921]$
    & $0.984\,[0.968,\,0.995]$ \\

\bottomrule
\end{tabular}
\end{table}

\begin{table}[!htbp]
\centering
\caption{Pairwise differences in test macro-F1 between frozen encoders. Values report point differences with 95\% confidence intervals estimated from 5,000 paired stratified bootstrap resamples of the held-out test set. Positive values favor the first encoder in each comparison.}
\label{tab:pairwise_differences}
\small
\setlength{\tabcolsep}{3pt}

\begin{tabular}{
@{}l
S[table-format=-1.3]@{\,[}S[table-format=-1.3]@{,\,}S[table-format=-1.3]@{]\quad}
S[table-format=-1.3]@{\,[}S[table-format=-1.3]@{,\,}S[table-format=-1.3]@{]\quad}
S[table-format=-1.3]@{\,[}S[table-format=-1.3]@{,\,}S[table-format=-1.3]@{]}
@{}}
\toprule
Dataset
& \multicolumn{3}{c}{\makecell{$\Delta$MedSAM2--SAM2}}
& \multicolumn{3}{c}{\makecell{$\Delta$MedSAM2--DINOv2}}
& \multicolumn{3}{c}{\makecell{$\Delta$SAM2--DINOv2}} \\
\midrule
Derm7pt
&  0.048 &  0.000 &  0.096
& -0.007 & -0.061 &  0.046
& -0.055 & -0.109 &  0.000 \\

HAM10000
&  0.046 &  0.004 &  0.090
& -0.008 & -0.048 &  0.034
& -0.054 & -0.101 & -0.008 \\

ISIC2020
&  0.002 & -0.005 &  0.008
& -0.053 & -0.063 & -0.043
& -0.054 & -0.064 & -0.044 \\

BCN20000
&  0.023 & -0.020 &  0.064
& -0.005 & -0.046 &  0.036
& -0.028 & -0.066 &  0.011 \\

DERM12345
& -0.023 & -0.128 &  0.087
& -0.048 & -0.115 &  0.054
& -0.025 & -0.142 &  0.108 \\

\midrule
Mean $\Delta$
& \multicolumn{3}{c}{0.019}
& \multicolumn{3}{c}{-0.024}
& \multicolumn{3}{c}{-0.043} \\
\bottomrule
\end{tabular}
\end{table}

We also inspected the hyperparameters obtained during validation-based model selection. The strongest DINOv2 runs did not rely on larger downstream heads: on HAM10000, and BCN20000, DINOv2 selected a one-layer MLP head. In contrast, the original SAM2 encoder selected deeper heads on BCN20000. MedSAM2 reduced this tendency, selecting shallower heads than vanilla SAM2 on these datasets. 

\section{Conclusion}
In this work, we evaluated whether SAM2 segmentation encoders can be reused as frozen image embeddings for skin-lesion classification. This question is motivated by the broader conjecture that segmentation foundation models learn general-purpose visual representations that may transfer beyond the task for which they were trained. We compared the original SAM2 encoder and the medically fine-tuned MedSAM2 encoder against DINOv2, a strong non-segmentation self-supervised visual baseline, across multiple dermatology datasets using a common frozen-encoder evaluation protocol. 

Our results show that SAM2-derived encoders contain diagnostic information that can be recovered by lightweight downstream classifiers. Medical adaptation generally improved the utility of these representations: MedSAM2 outperformed the original SAM2 encoder on most datasets by macro-F1. This suggests that medical-domain fine-tuning can make segmentation-derived features more useful for downstream dermatology classification, even when the encoder is kept frozen.

At the same time, the results do not provide evidence that SAM2-based segmentation encoders are superior frozen representations for skin-lesion diagnosis. DINOv2 obtained the highest macro-F1 point estimate on all evaluated datasets, although the margin over MedSAM2 was small in several cases and confidence intervals of this advantage (Table \ref{tab:pairwise_differences}) show uncertain results. Therefore, these results should be interpreted as evidence of partial transfer rather than definitive superiority of one representation family. The findings are consistent with the idea that segmentation encoders capture useful spatial and morphological information, while representation-oriented models may produce embeddings that are more directly separable for image-level diagnostic classification.

This study has several limitations. The experiments focus on frozen encoders with lightweight downstream heads, so they do not measure the performance achievable through full fine-tuning, adapter-based tuning, or end-to-end optimization. In addition, differences in dataset label taxonomies, class imbalance, acquisition conditions, and potential dataset overlap may affect direct comparisons across datasets.

Future work should evaluate whether segmentation-derived embeddings are useful beyond classification, including retrieval, clustering, nearest-neighbor analysis, and few-shot dermatology adaptation. It would also be valuable to study newer concept-aware segmentation models, such as SAM3, whose encoders may contain stronger semantic information than earlier SAM2-style models and could incorporate clinical information in a multimodal setup.

\section*{Data Availability}

All datasets used in this work are publicly available. The code and scripts required to reproduce the experiments are available in the project's repository.\footnote{\href{https://github.com/isicanonsubmission/Do-Segmentation-Encoders-Learn-Diagnostic-Representations}{https://github.com/isicanonsubmission/Do-Segmentation-Encoders-Learn-Diagnostic-Representations}}

    


%
%
%
\bibliographystyle{splncs04}
\bibliography{references}

\end{document}